\documentclass[conference]{IEEEtran}
\IEEEoverridecommandlockouts
\usepackage{cite}
\usepackage{booktabs}
\usepackage{amsmath,amssymb,amsfonts}
\usepackage[capitalize,noabbrev]{cleveref}

\usepackage{algorithmic}
\usepackage{graphicx}
\usepackage{textcomp}
\usepackage{xcolor}
\usepackage{multirow}

\def\BibTeX{{\rm B\kern-.05em{\sc i\kern-.025em b}\kern-.08em
    T\kern-.1667em\lower.7ex\hbox{E}\kern-.125emX}}
\begin{document}

\title{An Efficient and Streaming Audio Visual Active Speaker Detection System}
\author{
\IEEEauthorblockN{\begin{tabular}{c}Arnav Kundu, Yanzi Jin, Mohammad Sekhavat, Maxwell Horton, Danny Tormoen, Devang Naik\end{tabular}}
\IEEEauthorblockA{\{a\_kundu,yanzi\_jin,m\_sekhavat,mchorton,dtormoen,naik.d\}@apple.com \\Apple}}

\maketitle

\begin{abstract}
This paper delves into the challenging task of Active Speaker Detection (ASD), where the system needs to determine in real-time whether a person is speaking or not in a series of video frames. While previous works have made significant strides in improving network architectures and learning effective representations for ASD, a critical gap exists in the exploration of real-time system deployment. Existing models often suffer from high latency and memory usage, rendering them impractical for immediate applications. To bridge this gap, we present two scenarios that address the key challenges posed by real-time constraints. First, we introduce a method to limit the number of future context frames utilized by the ASD model. By doing so, we alleviate the need for processing the entire sequence of future frames before a decision is made, significantly reducing latency. Second, we propose a more stringent constraint that limits the total number of past frames the model can access during inference. This tackles the persistent memory issues associated with running streaming ASD systems. Beyond these theoretical frameworks, we conduct extensive experiments to validate our approach. Our results demonstrate that constrained transformer models can achieve performance comparable to or even better than state-of-the-art recurrent models, such as uni-directional GRUs, with a significantly reduced number of context frames. Moreover, we shed light on the temporal memory requirements of ASD systems, revealing that larger past context has a more profound impact on accuracy than future context. When profiling on a CPU we find that our efficient architecture is memory bound by the amount of past context it can use and that the compute cost is negligible as compared to the memory cost.
\end{abstract}


\section{Introduction}
\label{sec:intro}
With the shift from in-person to audio-visual online interactions, identifying active speakers in conversations has become crucial for effective communication and understanding. In multi-modal conversations, Active Speaker Detection (ASD) serves as a fundamental pre-processing module for speech-related tasks, including audio-visual speech recognition, speech separation, and speaker diarization. 

Working with ASD in real-world situations is challenging. It requires leveraging audio and visual information and understanding how they relate over time. To meet these needs, ASD models are designed to build audio-visual features and process them over long times to catch important timing information.

Most ASD frameworks \cite{alcazar2020active,alcazar2022end,talknet,light-asd,loconet} start with separate encoders that create embeddings for each modality and then use audio-visual fusion techniques to bring these different representations together so that they can be modeled jointly. The audio and visual encoders in these methods mostly use spatio-temporal CNNs and have some future context depending on the kernel size of the temporal CNN layers. Such an architectural choice introduces delay in the system because the encoders predictions are not aligned on the time axis (ie. latest output doesn correspond to the latest input frame). In our proposed solution, we make the encoders independent of future context making the output embeddings aligned in time.

The modality encoder embeddings are then fed into a combination step which can be as simple as concatenation or additon in the feature space \cite{light-asd} or more complicated like 
using cross-attention mechanism \cite{chakravarty2016cross,alcazar2020active}, which lets the visual streams attend to the audio streams at the same time.

After the two features are combined, self-attention \cite{alcazar2020active,alcazar2022end} or RNN-based \cite{light-asd} architectures are used to keep track of the person who is speaking the most clearly throughout the whole conversation. These methods take advantage of the fact that audio and visual information often provide complimentary signals. Additionally, there has been extensive recent work on making multi-modal representations stronger by learning good phonetic representations for the ASD task \cite{jung2024talknce}. 

However, to get more accurate models all these techniques use long temporal context (often the entire video) to make predictions for every frame \cite{alcazar2020active,alcazar2022end,talknet,light-asd,loconet}. This introduces a latency bottleneck since predictions for a given frame depend on all future frames. To this end, we propose a new model architecture that allows the model to have limited future context and the complete past context for a given frame to generate predictions. This way, we can maintain a fixed latency without significantly compromising accuracy. We perform 2 architectural changes to do this: we make the audio and visual encoders independent of future context and we make the fusion encoder constrained to future context during training.

Furthermore, to mitigate the high memory requirement caused by unlimited past context, we make the fusion encoder constrained to both past and future context. We then analyze the impact of various past and future contexts on the accuracy of the model to arrive to a configuration (memory and latency bounds) that is suitable for real-time applications while not compromising the accuracy significantly.
\begin{figure*}
    \centering
    \includegraphics[width=0.7\paperwidth]{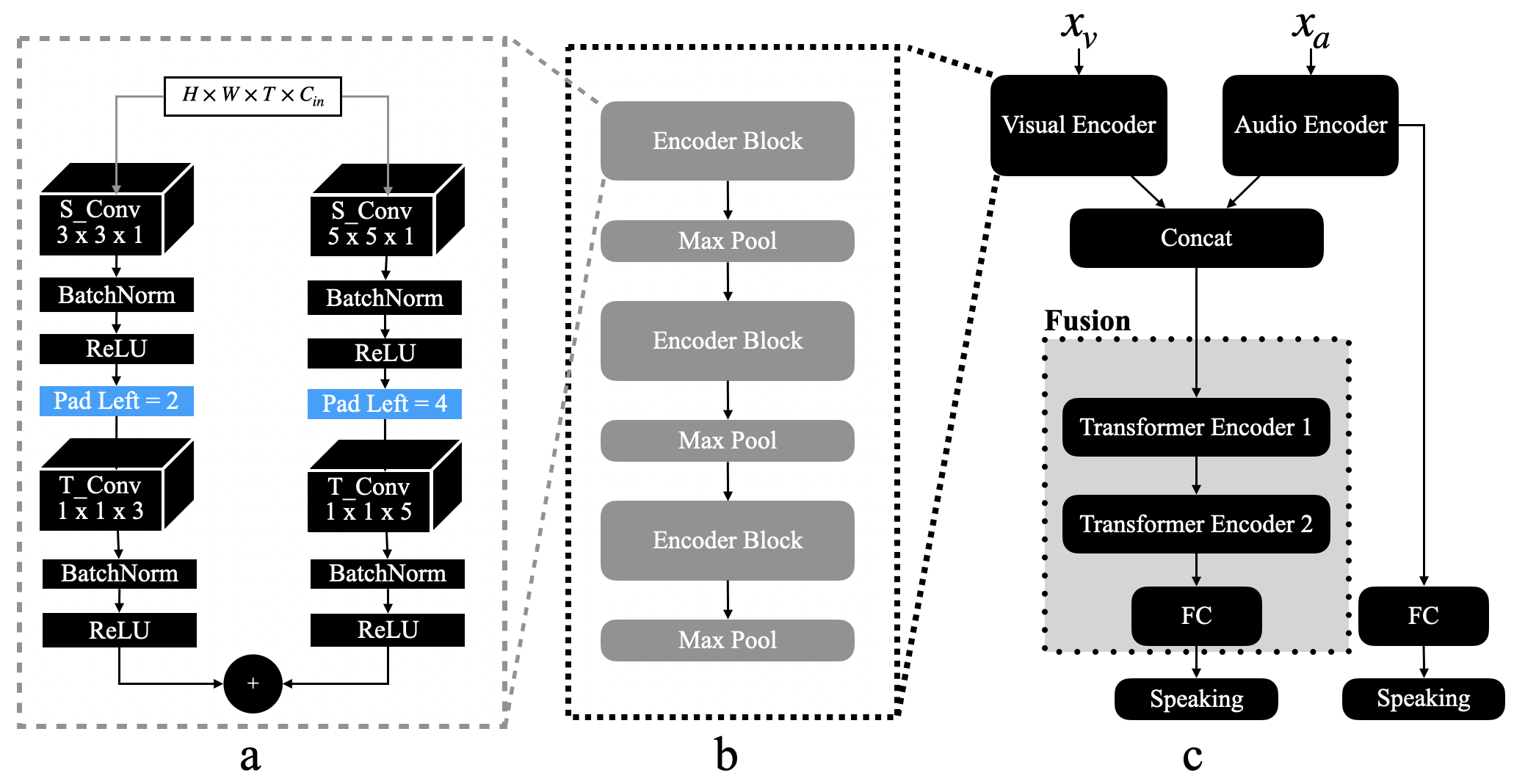}
    \caption{a: Visual Encoder Block (note the padding change in \textcolor{blue}{blue}), b: Visual / Audio Encoder, c: Full model architecture at train time}
    \label{fig:encoders}
\end{figure*}
\section{Related Works}
\label{sec:related}
This section outlines the general ASD framework commonly found in literature \cite{talknet,loconet, light-asd,alcazar2020active,alcazar2022end}. 

\textbf{Feature Extraction}:
The framework begins by processing the input video frames. Each face in the video is detected, cropped, stacked, and converted to normalized grayscale images, denoted as $x_v$. Correspondingly, the audio waveform is transformed into MFCC features, $x_a$. The extracted features are then fed into separate encoders to generate embeddings for the audio and visual modalities as follows:
\vspace{-0.5em}
\begin{equation}
\begin{split}
e_a = E_a(x_a)\\
e_v = E_v(x_v)
\end{split}
\end{equation}
\textbf{Fusion and temporal modeling}:
The audio-visual fusion module comes into play to integrate information between audio and visual cues. Recent methods such as TalkNet \cite{talknet} have introduced innovative approaches to this step. They utilize cross-attention layers, enabling effective alignment of the audio embedding with the visual information of the corresponding speaker. This leads to the generation of attention-weighted features, $f_a$ and $f_v$, which are then concatenated to form the fused audio-visual features $f_{av}$. This method is inspired from cross modal supervision used for voice activity detection using video. Other techniques like \cite{alcazar2020active, alcazar2022end, loconet} leverage the information from multiple speakers in the scene. Loconet  \cite{loconet} takes this inter-dependence further and leverages Long-term Intra-speaker Modeling (LIM) and Short-term Inter-speaker Modeling (SIM) in an interleaved manner. This makes the model aware of multiple people in the scene and handles edge cases like occluded faces. However, such architectures have very high parameter counts, therefore they might not be suitable for real-time applications. To adhere to real-time applications, we study Light-ASD \cite{light-asd} which uses a simple audio and visual backbone followed by a bi-directional GRU for active speaker detection. This model is trained end to end on the final task along with an auxiliary loss on the visual encoder to identify speaking labels using just the visual information. 

In all the above methods, the fusing consists of self-attention or bi-directional recurrent layers to model the temporal relationship of features. This makes these models contextually constrained to the entire video clip, i.e. to predict a label at time step 0 the model needs the inputs from the end of the video. This makes such models unsuitable for real-time applications. 
\section{Streaming ASD}
\label{sec:method}
For real-time applications, the models used for active speaker detection need to use only a restricted future context for a good runtime latency. We propose to solve this problem by enhancing the existing Light-ASD \cite{light-asd} architecture shown in \cref{fig:encoders}. Each video and audio encoder has two branches with different kernel sizes; a fusion model is applied to the concatenation of the audio/visual embedding. The future context can be introduced in two parts: the encoder and the fusion model. The audio and video encoders consist of 3-D CNN layers: spatial convolution S\_Conv  and temporal convolution T\_Conv) \cite{tran2018closer}. \cref{fig:encoders} represents the architecture of the visual encoder (the audio encoder is similar).  Light-ASD\cite{light-asd} encoders use equal padding (1 and 2 for kernel size of 3 and 5 in each branch respectively) for T\_Conv on both sides making them dependent on future context. To ensure streaming capabilities we ensure that the encoders do not rely on any future context and just the past context of a fixed receptive field. This has been done by modifying the temporal padding during training as shown in \textcolor{blue}{blue} in \cref{fig:encoders}. For the temporal convolutions (T\_Conv) of sizes 3 and 5, we left-pad the input by 2 and 4 respectively. In the audio encoder, the spatial convolution is replaced by a $3 \times 1 \times 1$ Conv3D layer.

Next in the fusion model, we address the problem of relying on long future context in our first iteration by replacing the bi-directional GRU in the Light-ASD architecture with a uni-directional GRU layer. As illustrated later in the experiments section, we found the accuracy of these models to be significantly worse than the baseline model. Given the success of transformers in replacing RNNs we replace the GRU layers from Light-ASD \cite{light-asd} with Transformer Layers. In transformers \cite{vaswani2017attention}, the output for a given time $T$ is a function of inputs from all timestamps. This relationship between input and output is established in the self-attention layer as illustrated in \cref{eq:attention}, where $Q_T$ represents the query vector corresponding to time $T$ and $K_t, V_t$ represent the key and value vectors used in self-attention from times $0 \le t \le T$. $\hat{T}$ represents the total time duration of the video. 
\begin{figure}
    \centering
    \includegraphics[width=1\linewidth]{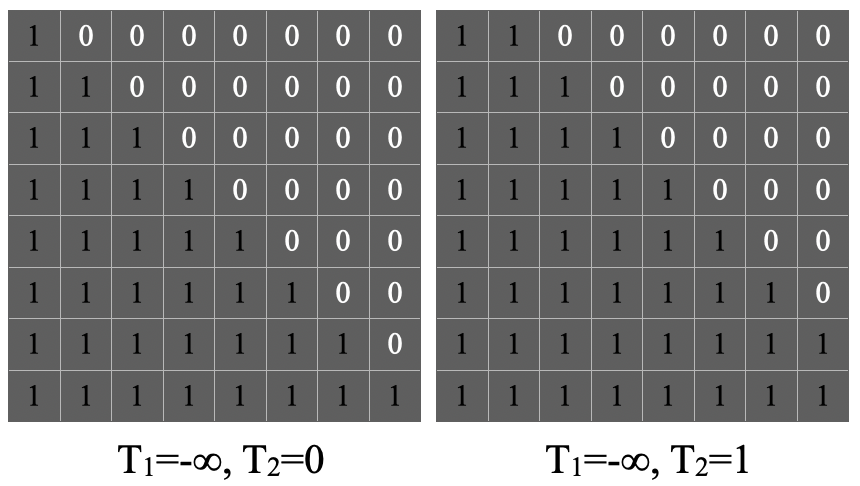}
    \vspace{-2em}
    \caption{Constrained mask for transformer encoder to limit the future context used by the model for predicting one label.}
    \label{fig:past_context_mask}
\end{figure}
\begin{figure}
    \centering
    \includegraphics[width=1\linewidth]{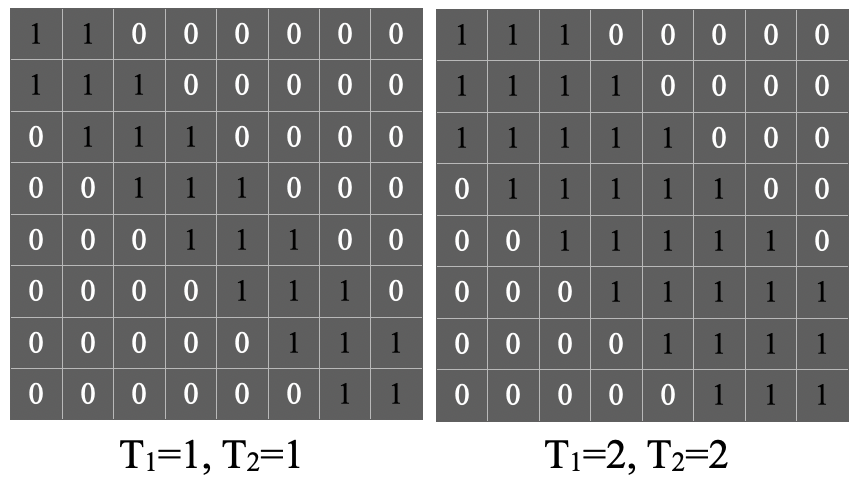}
    \vspace{-2em}
    \caption{Constrained mask for transformer encoder to limit the past and future context used by the model for predicting one label.}
    
    \label{fig:past_future_context_mask}
\end{figure}
\begin{equation}
\label{eq:attention}
    y_T = \Sigma_{t=0}^{\hat{T}}\text{softmax}(Q_T*K_t) *V_t
\end{equation}
To make sure that future context is not used in the computation of attention during training, we can modify \cref{eq:attention} to \cref{eq:constrained_future}.
\begin{equation}
\label{eq:constrained_future}
\begin{split}
    y_T &= \Sigma_{t=0}^{T}\text{softmax}(Q_T*K_t) *V_t\\
    &= \Sigma_{t=0}^{\hat{T}}\text{softmax}(Q_T*K_t * M) * V_t,\\
    \text{where } M&={1}_{\{t \leq T\}}.\\
\end{split}
\end{equation}
\begin{equation}
\label{eq:constrained_future_and_past}
\begin{split}
    y_T &= Q_T*\Sigma_{t=T-T_1}^{T-T_2}K_t *V_t\\
    &= Q_T*\Sigma_{t=0}^{\hat{T}}\text{softmax}(Q_T*K_t * M) * V_t,\\
    \text{where } M&={1}_{\{T-T_1 \leq t \leq T-T_2\}}.\\
\end{split}
\end{equation}

During training this uni-directional context is enforced by introducing a self-attention mask as shown in \cref{fig:past_context_mask} to ensure that the model cannot attend to any future inputs. This helps to use the same training graph and dataset without the need for writing custom code to split videos into clips to enforce uni-directional context.
This mask can also be modified to allow some future context by shifting the diagonal of this lower triangular matrix as shown in \cref{fig:past_context_mask}. The only problem with this approach would be using a huge KV-cache \cite{pope2023efficiently} to store the embeddings of the past inputs to compute the attention score with the current embeddings.

Therefore, to make the model memory efficient we can also limit the number of past frames the model can attend to by modifying the mask above to a new mask as shown in \cref{fig:past_future_context_mask} and described in \cref{eq:constrained_future_and_past}. This mask ensures that for predicting an output for time step $t$ inputs from only $t-T_1$ to $t+T_2$ are used, where $T_1$ is the number of past frames and $T_2$ is the number of future frames.

During inference, we do not need these masks and the model can be fed video and audio frames in a streaming fashion. The audio and video encoders being fully convolutional can process inputs frame by frame. The outputs from the encoders can be then fed into the transformer layers which maintain their own KV-cache \cite{pope2023efficiently} and use full self attention to predict the output corresponding to the $T^{th}$ frame for given input frames from $[X_{T-T_1},X_{T+T_2}]$.
\section{Experiments}
\label{sec:experiments}
\begin{table*}[htbp]
\begin{center}
\begin{tabular}{|c|c|c|c|c|c|c|c|c|}
\hline
\textbf{Model}& Encoder & Encoder & Fusion & Fusion &\textbf{mAP(\%)} & \textbf{Latency} & \textbf{Memory}\\
&Past (\#frames)&Future (\#frames)&Past (\#frames)& Future (\#frames)& & (ms)&\\
\hline
    Light-ASD \cite{light-asd} & $6$ & $6$ &  $\infty$ & $\infty$ & 94.06 & $\infty$ & $\infty$ \\
    Uni-directional GRU \cite{light-asd} & $6$ & $6$ &$\infty$ & 0 &  92.6 & 240 & 512KB \\
    \hline
    Bi-directional GRU & 12 & $0$ &$\infty$ & $\infty$ & 93.5 & $\infty$ & $\infty$ \\
    Uni-directional GRU & 12 & $0$ &$\infty$ & 0 &  91.9 & 0 & 512KB \\
    \hline
     & &  &$\infty$ & $\infty$ & \textbf{93.93} & $\infty$ & $\infty$ \\
     & &  &$\infty$ & 0 & 92.95 & 0 & $\infty$ \\
     & &  &$1$ & 1 & 91.13 & 40 & 512KB \\
    Transformer-Encoder & 12 & 0 &3 & 3 & 92.65 & 120 & 1.5MB \\
     & &  &6 & 12 & 92.93 & 480 & 3MB \\
     & &  &12 & 6 & 93.2 & 240 & 6MB \\
     & &  &32& 8 & 93.8 & 320 & 16MB \\
\hline
\end{tabular}
\vspace{1em}
\caption{Accuracy, Latency and Memory for combination of encoders and fusion models with different past and future contexts.}
\label{table:main_result}
\end{center}
\vspace{-2em}
\end{table*}
\subsection{Datasets}
The AVA-ActiveSpeaker dataset \cite{roth2020ava} is the go-to large-scale standard benchmark for active speaker detection. It features 262 Hollywood movies, 120 for training, 33 for validation, and 109 for testing (though the test set is withheld for the ActivityNet challenge). With over 5.3 million labeled face detection, speaking or nonspeaking, it contains many challenging scenarios like occlusions, low-resolution faces, low-quality audio, and tricky lighting. As is standard practice, we evaluate models on the validation set of AVA-ActiveSpeaker in our experiments. During training we augment the video frames by one rotation(-15 to 15 degrees), flipping or cropping (0 to 30\%) augmentations. Once an augmentation is selected it is applied to the entire video. For augmenting the audio frames we overlap the source audio with random audio sampled from the training data with SNR ratio of -5dB to 5dB.
\subsection{Implementation details}
The final architecture is built in PyTorch and trained on eight NVIDIA A100 GPUs (40GB). They are trained using the Adam optimizer \cite{kingma2014adam} for 60 epochs with a weight decay of $5e-4$. The learning rate for the transformer model is set to 0.003, which is then adjusted dynamically using a cosine learning rate scheduler \cite{loshchilov2016sgdr} throughout training which is generally a standard for most transformer models. For the GRU-based models, we use the same setting as the baseline Light-ASD setting ie. 0.001 learning rate with 5\% decay per epoch trained on a single GPU. We train our models from scratch on the AVA-ActiveSpeaker training dataset. Following the use of multi-task loss for AV-ASD \cite{light-asd,braga2022best} we use an auxiliary classification head from the video encoder to apply an auxiliary classification loss in addition to the classification loss from the fusion head.
\subsection{Evaluation criteria and metrics}

Following the common protocol suggested by previous works \cite{alcazar2020active, talknet, loconet, light-asd}, mean Average Precision (mAP) is used as an evaluation metric for the AVA- ActiveSpeaker validation set \cite{roth2020ava}. 

We divide our experiments into 2 parts: Encoders \textbf{with} and \textbf{without} future context, corresponding to the padding change in Figure \ref{fig:encoders}.
We have used Light-ASD \cite{light-asd} as our baseline model which uses a bi-directional GRU as its temporal model. As reported in the Light-ASD experiments the mAP for uni-directional GRU is significantly worse than that of the baseline model \cite{light-asd} with 240ms latency and minimal memory cost, shown as the top group in Table \ref{table:main_result}. Note that these models use encoders \textbf{with future context}.

Next, we train these models with encoders with \textbf{no future context} and find that the accuracy drops significantly. We hypothesize that transformers would be better suited as fusion models with such encoders and we validate it by training them as shown in \cref{table:main_result}. This acts as a good baseline for our further experiments with constrained transformers. Our results in \cref{table:main_result} indicate that a constrained transformer model outperforms the uni-directional GRU with 3 past and 3 future frames as context. Furthermore, we have shown that with 32 past and 8 future frames which translates to 16 MB of storage and 320 ms of latency in terms of wait time.

\begin{figure}
    \centering
    \includegraphics[width=1\linewidth]{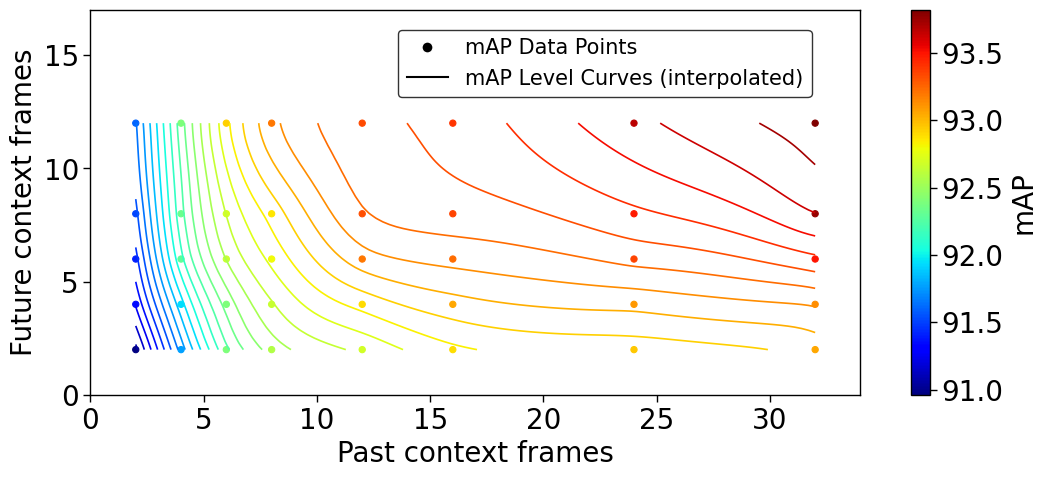}
    \vspace{-1.5em}
    \caption{Latency (future context frames) vs memory (past context frames) trade-off on accuracy (mAP\%). In this contour plot, the color of each data point (X,Y) indicates the mAP value corresponding to (memory=X, latency=Y). The mAP remains constant along each level curve, with the slope of these curves in different regions revealing which variable exerts greater influence. In areas where level curves are horizontal, future context primarily affects accuracy. Conversely, vertical level curves signify the dominant impact of past context. }
    \label{fig:latency_vs_mem_vs_map}
    \vspace{-1.5em}
\end{figure}

In \cref{fig:latency_vs_mem_vs_map}, we illustrate the impact of future context on the accuracy of our latency-constrained models. It is visible that beyond 6 future context frames the improvement in accuracy with more future context is marginal (color of the dots donot change for a fixed past context). We also investigate the impact of past context on the accuracy of the model as illustrated in \cref{fig:latency_vs_mem_vs_map}. It can be observed from the contour plot that past context has more impact on accuracy and only beyond a certain limit future context actually starts playing a role in the accuracy of the model.

We also profiled the execution time of our model on a Intel(R) Xeon(R) Platinum 8275CL CPU and found that the total runtime of our model is 32ms per frame with a given context of 32 past frames and 8 future frames. This makes the realtime latency of our model 352ms including the wait time for future frames. We found that compute time for self attention is just 5ms which allows us to avoid KV-caching and recompute all K,V for all frames. Therefore, we just store intermediate embeddings from the individual encoders making the total memory requirement 16MB (512KB per frame). 
\section{Conclusion}
\label{sec:conclusion}
We built a streaming and resource-constrained model using transformers for audio-visual active speaker detection to bring the gap to real-time application. We demonstrated that transformers are better suited as fusion models to capture temporal features. It has nearly state-of-the-art accuracy with a significantly lower latency and memory usage. 
We've done a comprehensive ablation study on the hyper-parameters that affect latency and memory usage in real-time. It showed including large future context and thereby increasing the latency of the model has very marginal gains in accuracy. Meanwhile having a large past context has more impact on the accuracy of the model than future context. Therefore, for streaming applications, it is important to have high memory availability for running ASD.
\newpage
\bibliographystyle{IEEEbib}
\bibliography{refs}
\vspace{12pt}

\end{document}